\documentclass[letterpaper, 10 pt, journal, twoside]{IEEEtran} 

\makeatletter
\let\NAT@parse\undefined
\makeatother

\usepackage{graphicx}
\usepackage{siunitx}
\usepackage[nolist,nohyperlinks]{acronym}
\usepackage{hyperref}
\usepackage{subcaption}
\usepackage{amsmath}
\usepackage{pifont}
\usepackage[flushleft]{threeparttable}
\usepackage{xcolor}
\usepackage{algorithm}
\usepackage{algpseudocode}
\usepackage{cite}

\newacro{GCP}[GCP]{Ground Control Point}
\acrodefplural{GCP}[GCPs]{Ground Control Points}
\newacro{TLS}[TLS]{Terrestrial Laser Scanner}

\newcommand{\cmark}{\ding{51}}%
\newcommand{\xmark}{\ding{55}}%

\title{Hilti SLAM Challenge 2023: Benchmarking Single + Multi-session SLAM across Sensor Constellations in Construction}

\author{Ashish~Devadas~Nair\textsuperscript{1},
        Julien~Kindle\textsuperscript{1},
        Plamen~Levchev\textsuperscript{1},
        and~Davide~Scaramuzza\textsuperscript{2}
\thanks{Manuscript received: April 12\textsuperscript{th}, 2024; Accepted June 10\textsuperscript{th}, 2024.}
\thanks{This paper was recommended for publication by Editor Javier Civera upon evaluation of the Associate Editor and Reviewers' comments.
This work was supported by Hilti Group and University of Zurich.} 
\thanks{\textsuperscript{1}Authors are with Corporate Research and Technology, Hilti~Group, Liechtenstein. {\tt\scriptsize <givenname>.<lastname>@hilti.com}}
\thanks{\textsuperscript{2}Dr. Davide Scaramuzza is with the Robotics and Perception Group, University~of~Zurich, Switzerland. {\tt\scriptsize sdavide@ifi.uzh.ch}}
\thanks{Digital Object Identifier (DOI): see top of this page.}
}

\begin{document}

\markboth{IEEE ROBOTICS AND AUTOMATION LETTERS. PREPRINT VERSION. ACCEPTED JUNE, 2024}
{Nair \MakeLowercase{\textit{et al.}}: Hilti SLAM Challenge 2023} 

\maketitle

\begin{abstract}
Simultaneous Localization and Mapping systems are a key enabler for positioning in both handheld and robotic applications.
The Hilti SLAM Challenges organized over the past years have been successful at benchmarking some of the world's best SLAM Systems with high accuracy. 
However, more capabilities of these systems are yet to be explored,
such as platform agnosticism across varying sensor suites and multi-session SLAM.
These factors indirectly serve as an indicator of robustness and ease of deployment in real-world applications. There exists no dataset plus benchmark combination publicly available, which considers these factors combined. The Hilti SLAM Challenge 2023 Dataset and Benchmark addresses this issue. Additionally, we propose a novel fiducial marker design for a pre-surveyed point on the ground to be observable from an off-the-shelf LiDAR mounted on a robot, and an algorithm to estimate its position at mm-level accuracy. Results from the challenge show an increase in overall participation, single-session SLAM systems getting increasingly accurate, successfully operating across varying sensor suites, but relatively few participants performing multi-session SLAM. Dataset URL: https://www.hilti-challenge.com/dataset-2023.html
\end{abstract}


\IEEEpeerreviewmaketitle

\section{Introduction}

\IEEEPARstart{T}{he} world of Simultaneous Localization and Mapping (SLAM) techniques has matured steadily as algorithms, sensors, computational capabilities, and simulation systems evolve. Most real-world applications of SLAM are becoming increasingly mission-critical with the advent of construction robotics, self-driving cars, and precision agriculture, among others. Both accuracy and robustness are imperative.

Multiple datasets and benchmarks have addressed various aspects of accuracy and robustness, as studied in Section \ref{sec:relatedWork}. However, there is no publicly available live benchmark that evaluates multi-session SLAM as of yet. The DARPA Subterranean Challenge~\cite{annurev:/content/journals/10.1146/annurev-control-062722-100728}, while expansive in its range of robots and environments covered, addressed multi-session SLAM benchmarking only to a limited extent where few chosen teams collected their data with multiple robots and were ranked based on artifact detection and positioning accuracy. A publicly accessible dataset plus benchmark would advance multi-session and multi-device SLAM research worldwide.

\begin{figure}[ht]
  \hspace*{5mm}
  \begin{subfigure}[b]{0.2\textwidth} 
    \includegraphics[width=\textwidth]{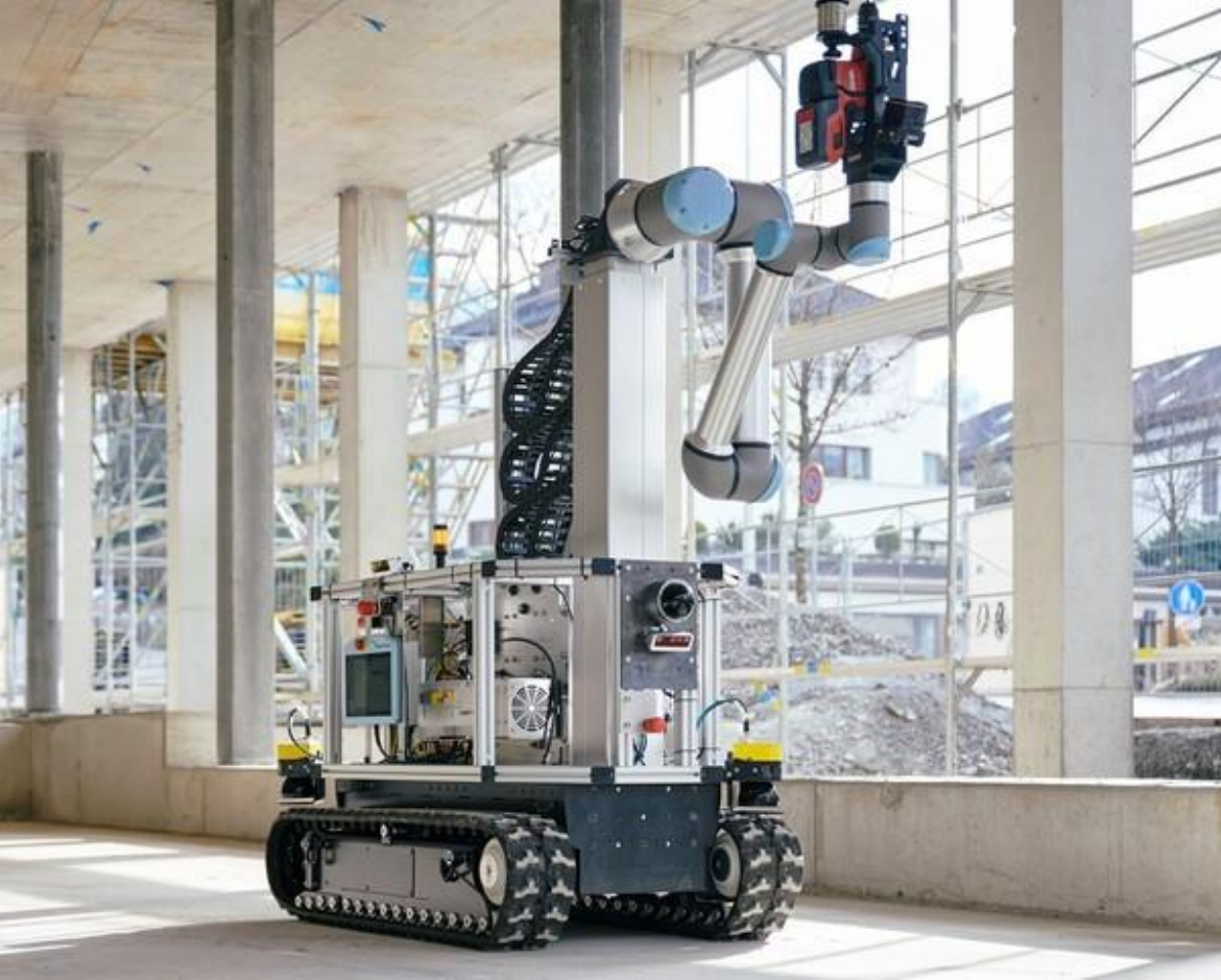}
    \captionsetup{justification=centering}
    \caption{Trailblazer - drilling robot prototype.}
    \label{fig:trailblazer}
    \vspace*{2.5mm}
    \includegraphics[width=\textwidth]{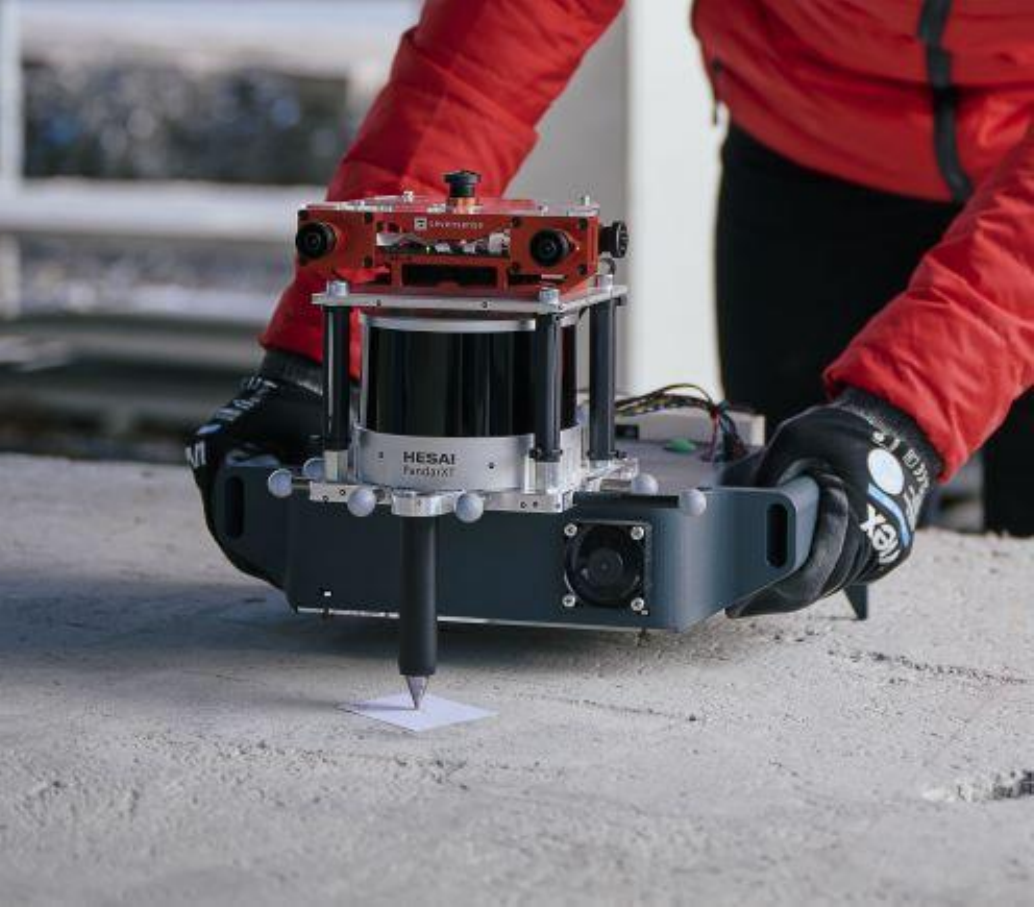}
    \captionsetup{justification=centering}
    \caption{Phasma - Hand-held scanner prototype.}
    \label{fig:phasma}
  \end{subfigure}
  \hspace*{2mm}
  \begin{subfigure}[b]{0.2\textwidth} 
    \includegraphics[width=0.95\textwidth]{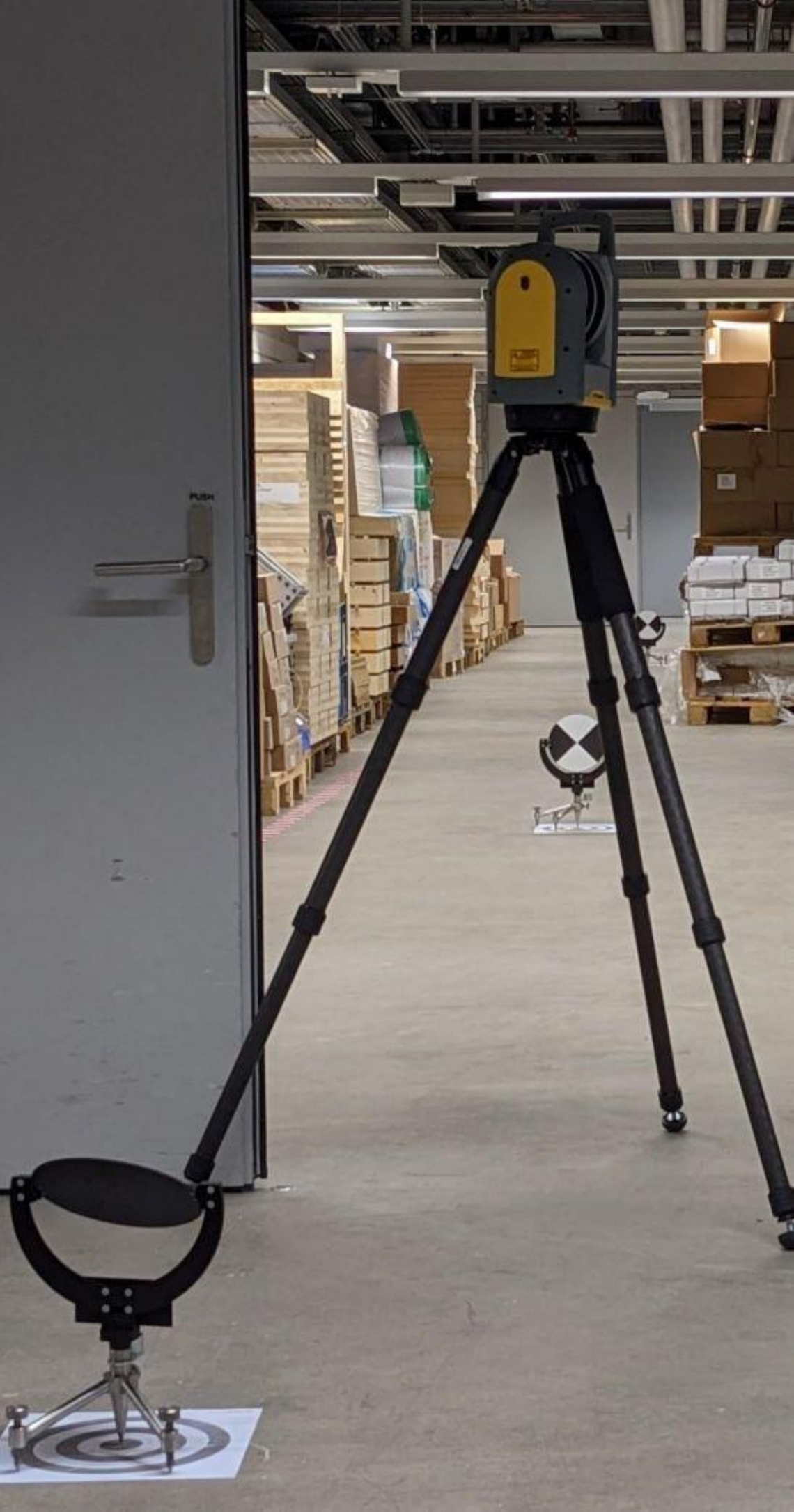}
    \captionsetup{justification=centering}
    \caption{Trimble X7 - Terrestrial Laser Scanner, used for Ground Control Point (GCP) extraction.}
    \label{fig:TLS-x7}
  \end{subfigure}
  \captionsetup{justification=centering}
  \caption{Our SLAM data acquisition devices (a,b); and Ground Truth extraction approach (c).}
    \label{fig:hw_overview}
\end{figure}

The Hilti SLAM Challenge 2022~\cite{Zhang_2023} proposed a system where SLAM trajectories were compared with surveyed points in a scene, where during data collection for SLAM, a human operator physically placed a handheld device at survey points. There exists a known rigid body transform between the sensors' reference frame and the spike that made physical contact with the survey point, as seen in Figure ~\ref{fig:phasma}. However, this approach of physical contact is not extensible for mobile robot-based SLAM evaluation, since it may not have a human operator to help with fine positioning of a spike. A solution is needed to this, where the robot can use its sensors to detect and estimate such a survey point’s position accurately. Furthermore, most SLAM benchmarks today use the IMU reference frame to evaluate trajectories, hence the extrinsic transformation of this estimated survey point from the measuring sensor's reference frame to the IMU reference frame needs to be accurate, for accurate Absolute Trajectory Error (ATE) calculation~\cite{grupp2017evo,8593941}. Allowing SLAM benchmark submissions to include custom extrinsics in ATE evaluation would address inaccuracies introduced by the calibration data provided with a SLAM dataset.

We also draw into context the field of construction, where SLAM-based systems operate in various site conditions, need to restart or continue from where a task was previously left incomplete, expand on a previous map of the environment, and possibly collaborate with another device or robot. Given these needs, we designed the Hilti SLAM Challenge 2023 to deliver:
\begin{itemize}
    \item A robot mounted sensor suite: A Construction Robot platform (Figure \ref{fig:trailblazer}) with four OAK-D cameras, a Robosense B-Pearl hemisphere Lidar, and an XSens MTi-670 IMU. This is in addition to the handheld Phasma (Figure \ref{fig:phasma}) prototype used in the Hilti SLAM Challenge 2022~\cite{Zhang_2023}.
    \item A lidar-observable fiducial marker and its center position estimation algorithm. A survey point on the ground with such a fiducial is here referred to as a Ground Control Point (GCP), as seen in Figures \ref{fig:TLS-x7} and \ref{fig:tb_sensors}.
    \item A multi-device, single and multi-session SLAM Dataset comprising of three locations.
    \item Revised Benchmarking and Evaluation System that includes options for single session SLAM for both platforms, and multi-session SLAM evaluation for each.
    \item The option for participants to include their own calibration files in their SLAM trajectory submissions.
\end{itemize}

\section{Related Work} \label{sec:relatedWork}
Increased focus in recent years on research in autonomous driving and mobile robotics has driven the rise of low-cost lidars. This has enabled more research groups worldwide to produce SLAM datasets and benchmarks. 
Our previous Hilti SLAM Challenge iterations from 2021~\cite{Helmberger_2022} and 2022~\cite{Zhang_2023} have set the bar high in pushing the limits of SLAM system capabilities in construction site. Both consisted of a human-operated Lidar, inertial, and visual sensor setup, with evaluations of ATE. The ground truth (here referred to as GT) methodologies have varied over the years. The 2021 Challenge witnessed the use of a Hilti PLT300 total station or a motion capture system for dense ground-truth pose estimation but was limited in range due to line-of-sight operation---this prevented exploration of complex areas. To deal with this, the 2022 Challenge made use of a Z+F Imager 5016 Terrestrial Laser Scanner (here referred to as TLS) as a ground-truth map source, along with surveyed control points extracted from the map as position ground truth. This enabled sub-centimeter sparse GT accuracy even in largely varying scenes, a methodology we chose to adopt.

While addressing SLAM benchmarking across multiple platforms and sessions, the closest fit is the ICCV SLAM Challenge 2023 (October 2023), which encompassed the real-world SubT-MRS~\cite{zhao2023subtmrs} and the simulated TartanAir~\cite{wang2020tartanair} datasets to provide dense ATE and Relative Pose Error RPE evaluations in a variety of indoor, outdoor and subterranean scenes. The SubT-MRS dataset is notable due to its large span of sensors and platforms. The ICCV SLAM Challenge benchmark uses a TLS for GT map extraction followed by a lidar frame to map registration for GT trajectory traces, similar to the Hilti SLAM Challenge 2022's supplementary dense 6DoF traces. However, its ground truth trajectory accuracy used for evaluation is in the +/- 10cm range ~\cite{zhao2023subtmrs}, due to the lidar frame scan to map alignment error, and is susceptible to lidar degeneracy cases. 

We appreciate that the ICCV SLAM Challenge has adopted benchmarking tracks similar to the Hilti SLAM Challenge series. However, it does not evaluate multi-session SLAM, and they do not provide participants the option to provide their own extrinsic calibration for ATE evaluation.

Observing construction-related SLAM datasets makes it worth mentioning ConSLAM~\cite{trzeciak2023conslam, trzeciak2023conslamExtension} and ConPR~\cite{lee2023conpr}. ConSLAM is a handheld SLAM dataset consisting of lidar, inertial, RGB, and NIR camera data. Its ground truth is a similar scan-to-map-based pose extraction system as seen before, causing similar limitations with overall ATE accuracy. The interesting feature is that sessions in the dataset were collected across various stages of construction, showing potential towards long-term SLAM. ConPR does the same on the handheld scanning aspect but with a perspective lidar instead of a scanning lidar, and also includes GPS measurements. Ground truth trajectories are a fusion of FastLIO2~\cite{xu2021fastlio2} and GPS. However, for most construction-related applications that leverage handheld systems or robots, consistently high-accuracy positioning across various conditions is the key driver of better quality control and human-equivalent task performance for robots in real-world conditions, emphasizing the need for TLS-based GT.

\section{Hardware}
\begin{figure}[h]
  \begin{subfigure}[b]{0.24\textwidth}
    \begin{center}
    \includegraphics[width=0.755\textwidth]{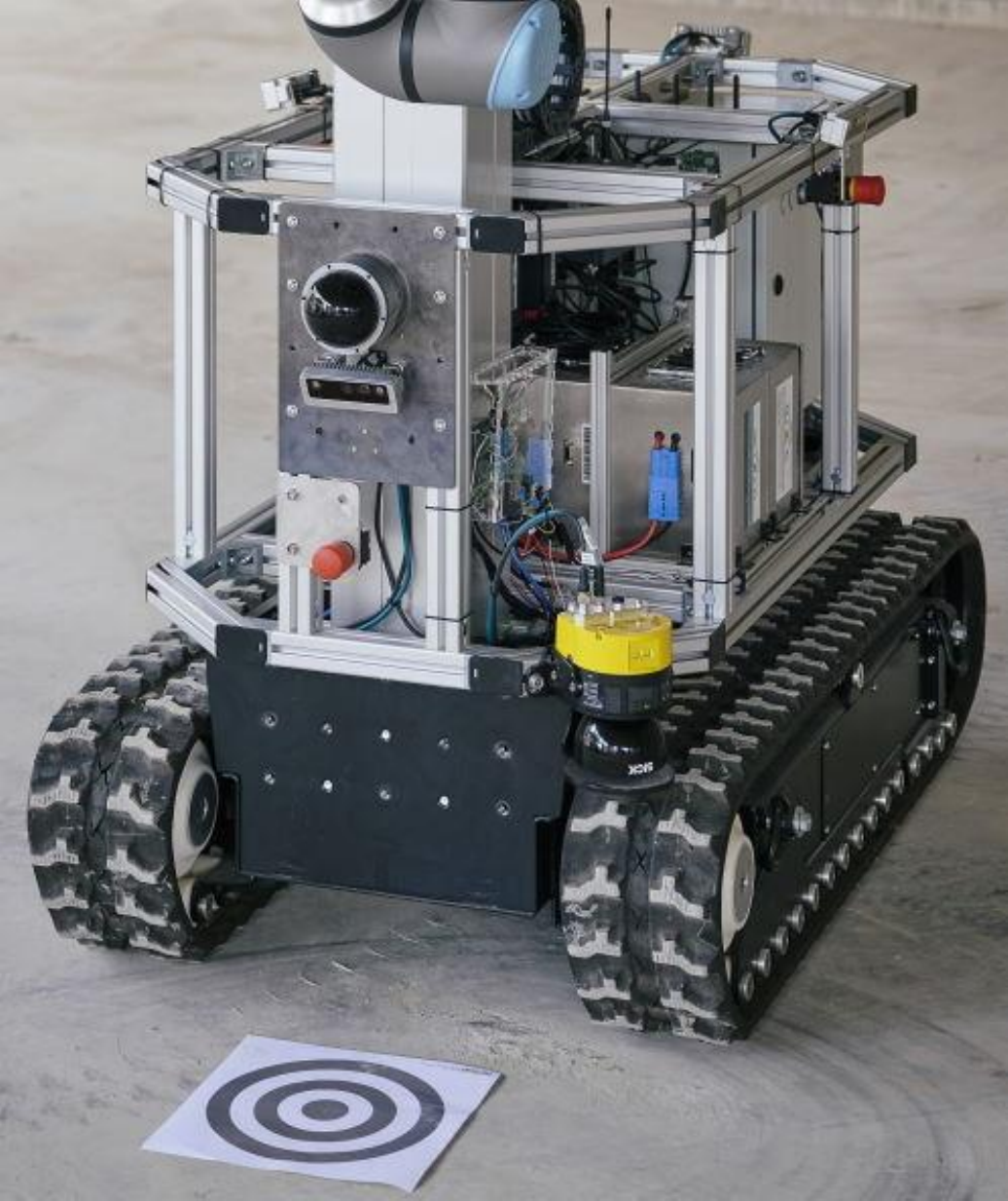} 
    \end{center}
    \captionsetup{justification=centering}
    \caption{It is equipped with a front facing hemisphere lidar, 4x stereo cameras, and an IMU.}
    \label{fig:tb_sensors}
  \end{subfigure}
  \begin{subfigure}[b]{0.24\textwidth}
    \begin{center}
    \includegraphics[width=0.8\textwidth]{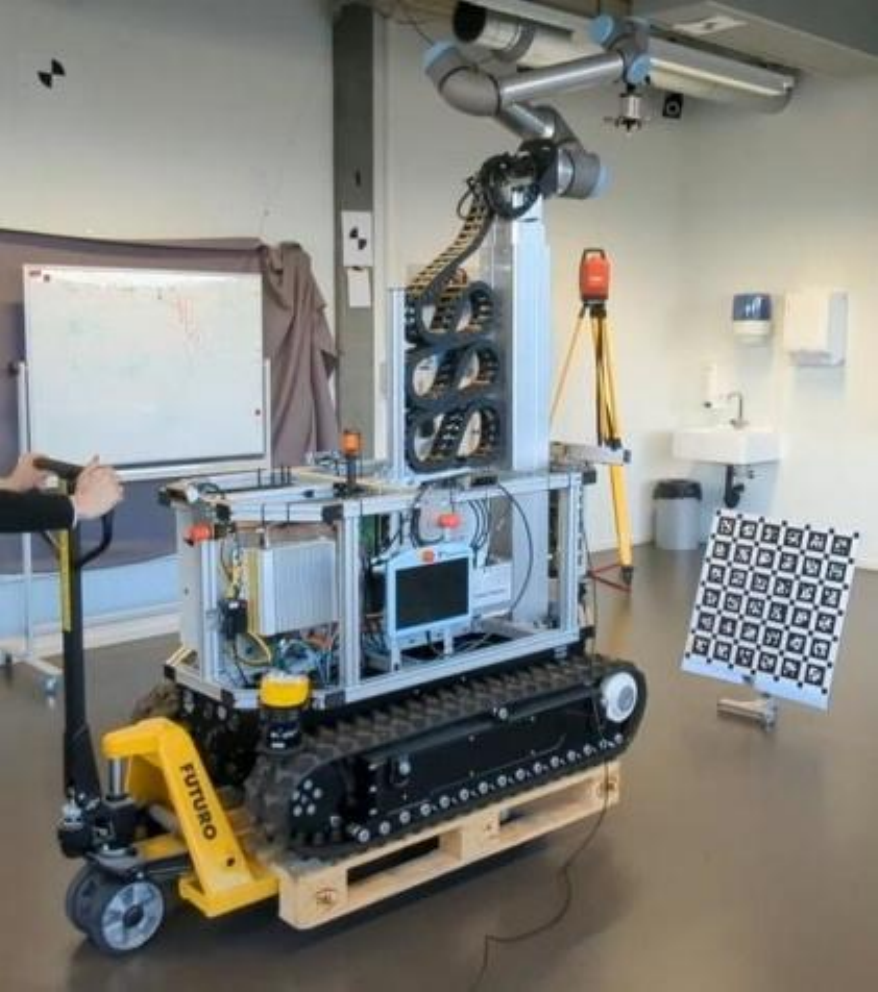}
    \end{center}
    \captionsetup{justification=centering}
    \caption{It was moved with a pallet jack during its calibration sequence.\\}
    \label{fig:tb_calibration}
  \end{subfigure}
  \captionsetup{justification=centering}
  \caption{Trailblazer's sensor suite (a) \\and calibration data collection procedure (b).}
    \label{fig:tb_overview}
\end{figure}

In comparison to our previous SLAM Challenge~\cite{Zhang_2023} in which we recorded the datasets exclusively with our hand-held device called Phasma\footnote{\tiny\url{https://github.com/Hilti-Research/hilti-slam-challenge-2023/blob/main/documentation/hardware/Handheld.md}} (Figure~\ref{fig:phasma}), this year's dataset features additionally recorded trajectories with our robot prototype called Trailblazer (Figure~\ref{fig:tb_overview}), a \SI{700}{\kilo\gram} construction robot. In contrast to Phasma, the datasets recorded with Trailblazer include robot-specific behaviors such as strong vibrations when turning, mostly constrained motion to a plane and large extrinsics between sensors. Since we chose to maintain the same hardware and calibration routines for the Phasma prototype as the Hilti SLAM Challenge 2022 \cite{Zhang_2023}, this section focuses primarily on the hardware setup of Trailblazer.

Trailblazer consists of a horizontally mounted Robosense Bpearl hemispherical LiDAR (+/-3cm noise) with 32-rays, four Luxonis OAK-D stereo camera pairs all around the system and slightly inclined towards the floor and an XSens MTi-670 IMU. The LiDAR, one stereo camera and the IMU are mounted on a steel plate to minimize changes in extrinsics. The LiDAR is synchronized to the PC directly using Precision Time Protocol (PTP) while the IMU and the cameras use a PTP-to-trigger board which triggers the start of exposure of all cameras. Each stereo camera pair independently adjusts exposure based on the observed scene.
The calibration sequences were intentionally recorded with the full assembly to reflect challenges arising from a heavy construction-site robot with an emphasis on in-field calibration. This means that for all calibration sequences, we did not unmount individual sensors or sub-assemblies of the system and only used tools which could easily be used at a construction site.

A new challenge arising from using robots in the context of the Hilti SLAM Challenge is the recording of \acp{GCP} in robot sequences: unlike Phasma, positioning a tip mounted to a tracked robot on a \ac{GCP} with mm-accuracy is not easily achieved. In the following sub-sections, we describe our approach to sensor calibration as well as the registration of \acp{GCP}.

\subsection{IMU Calibration}
For the calibration of the IMU, we recorded the IMU measurements at standstill, sampling at \SI{200}{\hertz} overnight for approximately \SI{11.5}{\hour}. We utilized Allan Variance estimation~\cite{gao2018imu_utils, UCAM-CL-TR-696} as the default calibration method to estimate the noise densities and random walks of the accelerometer and gyroscope. During this recording, we kept all systems active (LiDAR, robot fans, etc.) to capture these steady-state vibrations in the calibration data.

\subsection{Stereo-Camera Calibration}\label{sec:cam_calib}
For the calibration of the stereo camera pairs, we moved a 6x6 AprilTag grid, measuring \SI{60}{\centi\meter}, for approximately \SI{60}{\second} in front of each sensor. We fixed the exposure time of the sensors to \SI{1}{\milli\second} and set the ISO to \SI{200}{}. The provided default calibration, which includes the intrinsics of each camera and the extrinsics between stereo camera pairs, was created by running Kalibr~\cite{6696514} on each individual stereo camera dataset.

\subsection{Extrinsic Calibration}
To sufficiently excite all degrees of freedom of Trailblazer, we placed the robot on a pallet and moved it around with a hand pallet jack. This method allowed us to not only excite the degrees of freedom in the plane ($x$, $y$, \emph{yaw}) but also to induce \emph{roll} and \emph{pitch} by shaking the system about the $x$ and $y$ axes. Additionally, it enabled limited motion in the $z$ axis (under \SI{10}{\centi\meter}).
In the calibration dataset, we used six 6x6 AprilTag grids, each measuring \SI{60}{\centi\meter}, arranged in a hemisphere with the boards at different heights. The center board was equipped with high reflective tape for easy detection in the point cloud. The dataset, which covered the excitation of all degrees of freedom, lasted approximately \SI{240}{\second}.
To create the default extrinsic calibration, we used MultiCal~\cite{9982031} and developed an adapted version\footnote{\tiny\url{https://github.com/Hilti-Research/multical}} that allows for the freezing of the extrinsics of stereo camera pairs. This adaptation was motivated by our greater confidence in the results obtained from the intrinsic calibration of the cameras (Section~\ref{sec:cam_calib}), as we were able to move the AprilTag grids across the full field of view of each camera.

\subsection{Registering of Ground Control Points}
The accuracy evaluation of submitted SLAM trajectories requires to relate the pose of the robot to ground truth measurements from the \ac{TLS}. On Phasma, this is achieved by accurately placing a steel tip calibrated w.r.t. all sensors on a measured-in \acp{GCP}. On Trailblazer, this method is not feasible as positioning such a spike with high accuracy would take too long. Instead, we want to be able to directly measure a \ac{GCP} with sensor data. In previous Hilti SLAM Challenges, we have observed that the accuracy of the global estimate mainly results from the LiDAR measurements - therefore, to reduce potential errors introduced by the extrinsic calibration between the LiDAR and a camera, we decided to detect the \acp{GCP} from LiDAR data directly. 

\begin{figure}[h]
  \begin{subfigure}[b]{0.23\textwidth}
    \includegraphics[width=\textwidth]{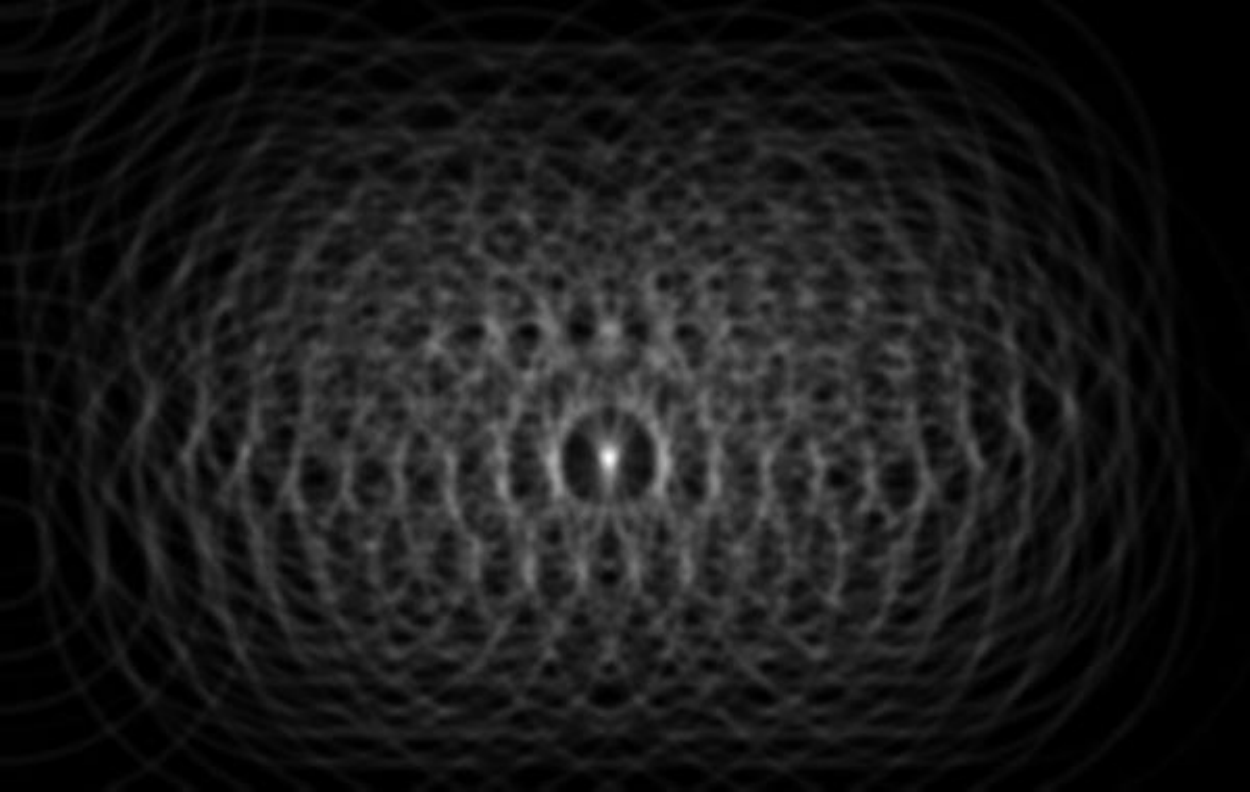}
    \caption{Circular Hough Space representation of a GCP detection.\\}
    \label{fig:gcp_hough_a}
    \vspace*{3mm}
    \includegraphics[width=\textwidth]{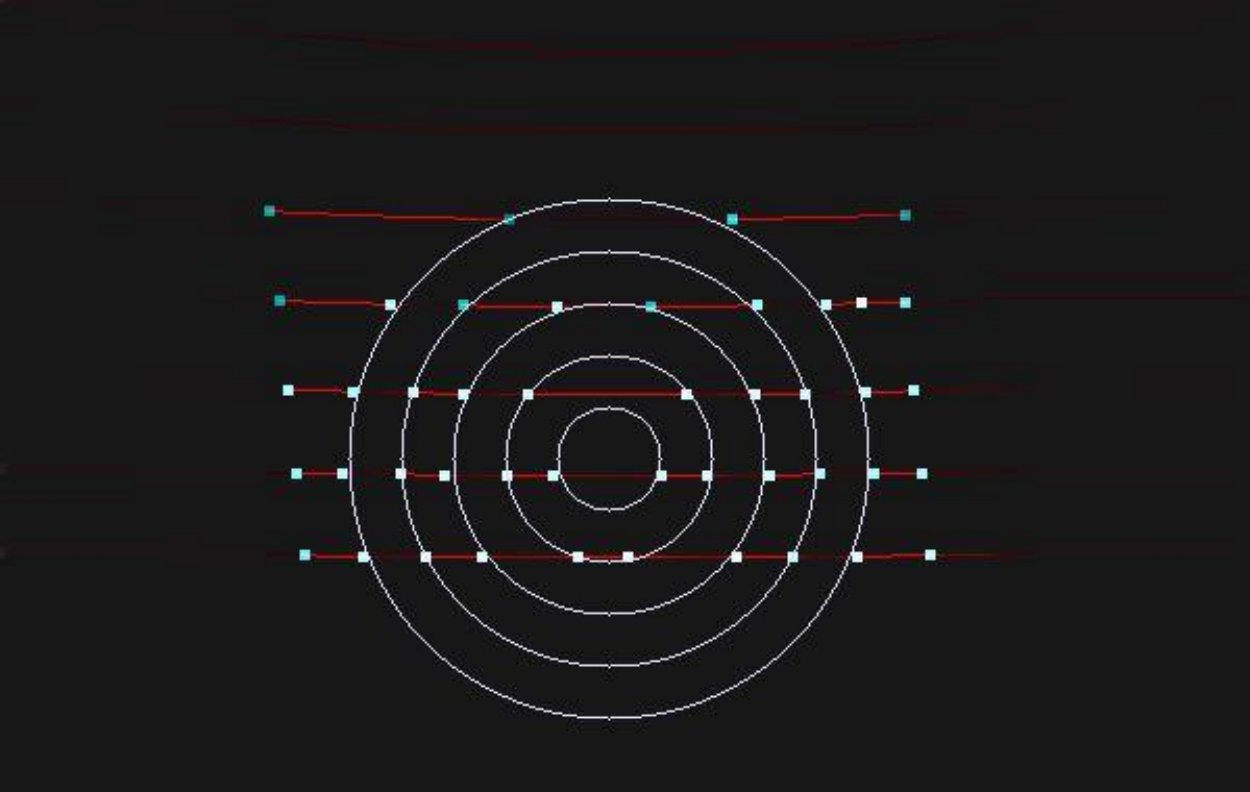}
    \caption{Detected edges of LiDAR intensity, and the resulting most likely fit of the GCP overlayed.}
    \label{fig:gcp_hough_b}
  \end{subfigure}
  \hfill
  \begin{subfigure}[b]{0.23\textwidth}
    \includegraphics[width=0.99\textwidth]{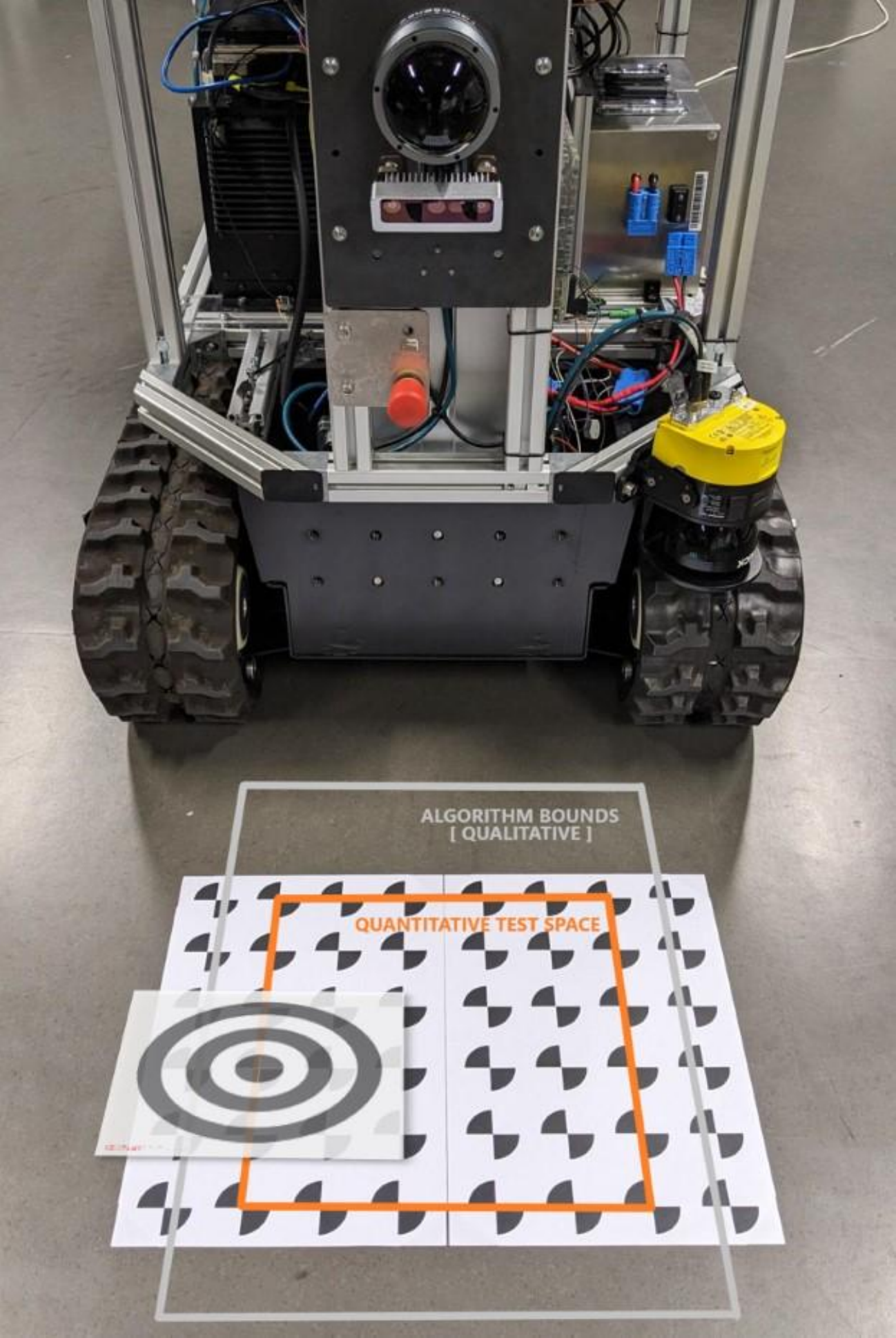}
    \caption{Pre-surveyed Test Grid, annotated with quantitative (orange), qualitative (grey) test bounds, and an overlayed GCP target.}
    \label{fig:gcp_grid}
  \end{subfigure}
  \caption{GCP detector visualizations (a,b) and test setup (c).}
    \label{fig:gcp_algo_testing}
\end{figure}

LiDARTag~\cite{HuangLiDARTag2020} is a LiDAR detector framework which allows to localize apriltags directly from LiDAR data. Due to the sparsity of the Robosense Bpearl, however, the size of the targets would need to be disproportionately large. Additionally, as LiDARTag searches for spatial discontinuities, we would not be able to place the targets on the floor to use them for both Trailblazer and Phasma. Therefore, we propose our own \ac{GCP} detector for detecting circular targets as seen in Figure~\ref{fig:gcp_grid}. The structure of the \ac{GCP} detector is outlined in Algorithm~\ref{alg:gcp_detector}. We project the region-of-interest of multiple LiDAR scans to the fitted ground plane. For each scan and ring, we detect large changes in intensity using a 1D Canny Edge detector (Figure ~\ref{fig:gcp_hough_b}) and add these as votes to the Hough Space as multiple circles with radii from the \ac{GCP} (Figure~\ref{fig:gcp_hough_a}). Finally, we smoothen the Hough Space and transform the most likely candidate back to 3D space.

Note that in the Bpearl lidar, triggering is implemented on a time basis instead of encoder values that always measure at the same angles with every pass. Due to slight variability in the rate of rotation the resulting measurements also vary, leading to higher resolution in edge detection with successive scans at the same position. Furthermore, given the position and orientation of the lidar, a GCP located within 90 cm from the device will always have at least 3 scan lines passing through it, which yields favorable sub-centimeter accuracy.

We evaluated the relative accuracy of our approach by detecting the center of the \ac{GCP} placed on multiple locations of a 6x6 grid, $\tilde{t}_{\text{lidar},p_i}$ as seen in Figure \ref{fig:gcp_grid}, which was pre-surveyed using a Trimble X7 TLS, details in Section \ref{sec:GT}. Then, the transform between the LiDAR and the grid $T_\text{grid,lidar}$ is determined using the Kabsch Algorithm~\cite{Kabsch1976}. The error is then computed as:
\begin{equation}
    e_i = t_{\text{grid},p_i} - T_\text{grid,lidar}\,\, \tilde{t}_{\text{lidar}, p_i}
\end{equation}

\begin{algorithm}
\small{
\caption{LiDAR \ac{GCP} Detection Algorithm}\label{alg:gcp_detector}
\begin{flushleft}
\textbf{Input:} Set of cropped LiDAR scans $\mathbf{S}_\text{lidar}$\\
\textbf{Output:} Estimated position of \ac{GCP} center
\begin{algorithmic}[1] 
\State $T_\text{plane,lidar} \gets \text{fitPlane}(\mathbf{S}_\text{lidar})$ \Comment{Fit plane to points}
\State $\mathbf{U}_\text{plane} \gets \text{proj}(T_\text{plane,lidar},\mathbf{S}_\text{lidar})$ \Comment{Project points to plane}
\State $H \gets 0$ \Comment{Initialize empty Hough Space}
\For{$U \in \mathbf{U}_\text{plane}$} \Comment{Iterate over all scans}
    \State $C \gets 0$ \Comment{Initialize empty set of corners}
    \For{$r \in U$} \Comment{Iterate over all rings}
        \State $C \mathrel{+}= \text{detectEdges}(r)$ \Comment{1D Canny edge}
    \EndFor
    \State $H \mathrel{+}= \text{houghSpace}(C)$ \Comment{Add Hough Space votes}
\EndFor
\State $H \gets \text{gaussianBlur}(H)$ \Comment{Smoothen Hough Space}
\State $v_\text{max} \gets \text{argmax}(H)$ \Comment{Find most likely position in plane}
\State \textbf{return} $T_\text{plane,lidar}^{-1}\,\,v_\text{max}$ \Comment{Transform to 3D Space}
\end{algorithmic}
\end{flushleft}
}
\end{algorithm}

\begin{figure}[h]
    \centering
    \includegraphics[width=0.47\textwidth]{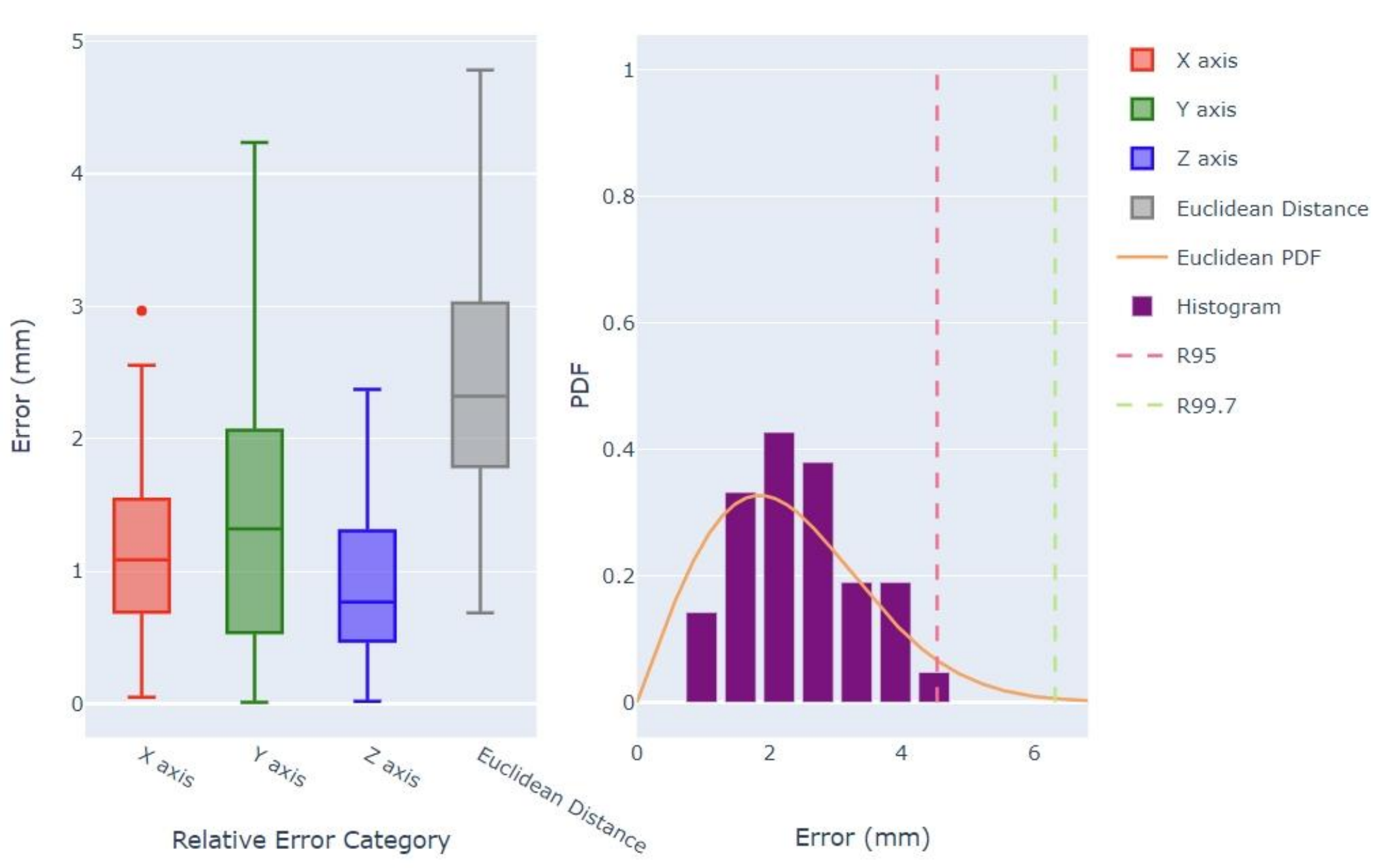}
    \captionsetup{justification=centering}
    \caption{GCP Estimation Accuracy Evaluation. \\Left: GCP Estimation Relative Error box plots. \\Right: Rayleigh Distribution of 3DoF Euclidean Error.}
    \label{fig:gcp-acc-plot}
\end{figure}

The evaluation results are summarized in Figure~\ref{fig:gcp-acc-plot}. We document errors in individual axes, as well as the 3-DoF euclidean distance error. 3-DoF Euclidean distance error of the estimated points reflects the combined effect of errors from all three axes, with a median of \SI{2.3}{\milli\meter}, and a max of \SI{4.7}{\milli\meter}, compared to the surveyed points. A fitted Rayleigh distribution on the normalized histogram yields an R95 of \SI{4.54}{\milli\meter} and an R99.7 value of \SI{6.32}{\milli\meter}.

Note that our evaluation approach only captures relative accuracy, as determining the physical 6-DoF location of the LiDAR frame is a challenge in itself. It evaluates how accurately we can determine the position of multiple \ac{GCP} relative to each other but does not account for absolute effects such as the range-dependent bias of the sensor. For the evaluation of the SLAM Challenge datasets, however, the impact of absolute effects should be negligible as all \ac{GCP}s were placed on the same plane that the robot stood on.

\begin{figure*}
    \centering
    \begin{subfigure}[t]{0.54\linewidth} 
      \centering
      \includegraphics[width=\linewidth]{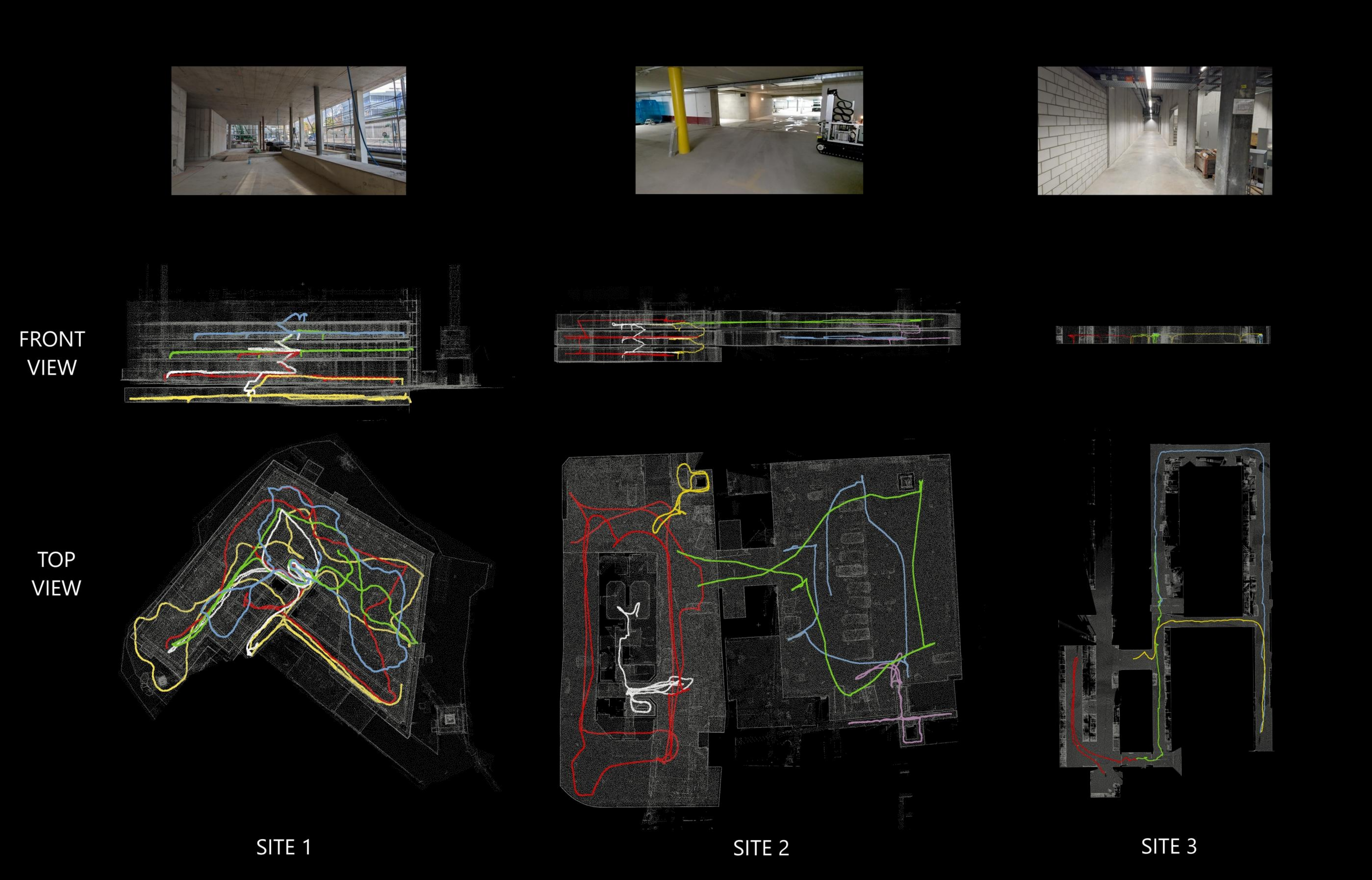}
      \captionsetup{justification=centering}
      \caption{Locations: Images, Downsampled TLS Scans, and globally aligned FastLio2/AdaLio based Trajectories.}
      \label{fig:Site-X}
    \end{subfigure}%
    \hfill 
    \begin{subfigure}[t]{0.36\linewidth} 
      \centering
      \includegraphics[width=\linewidth]{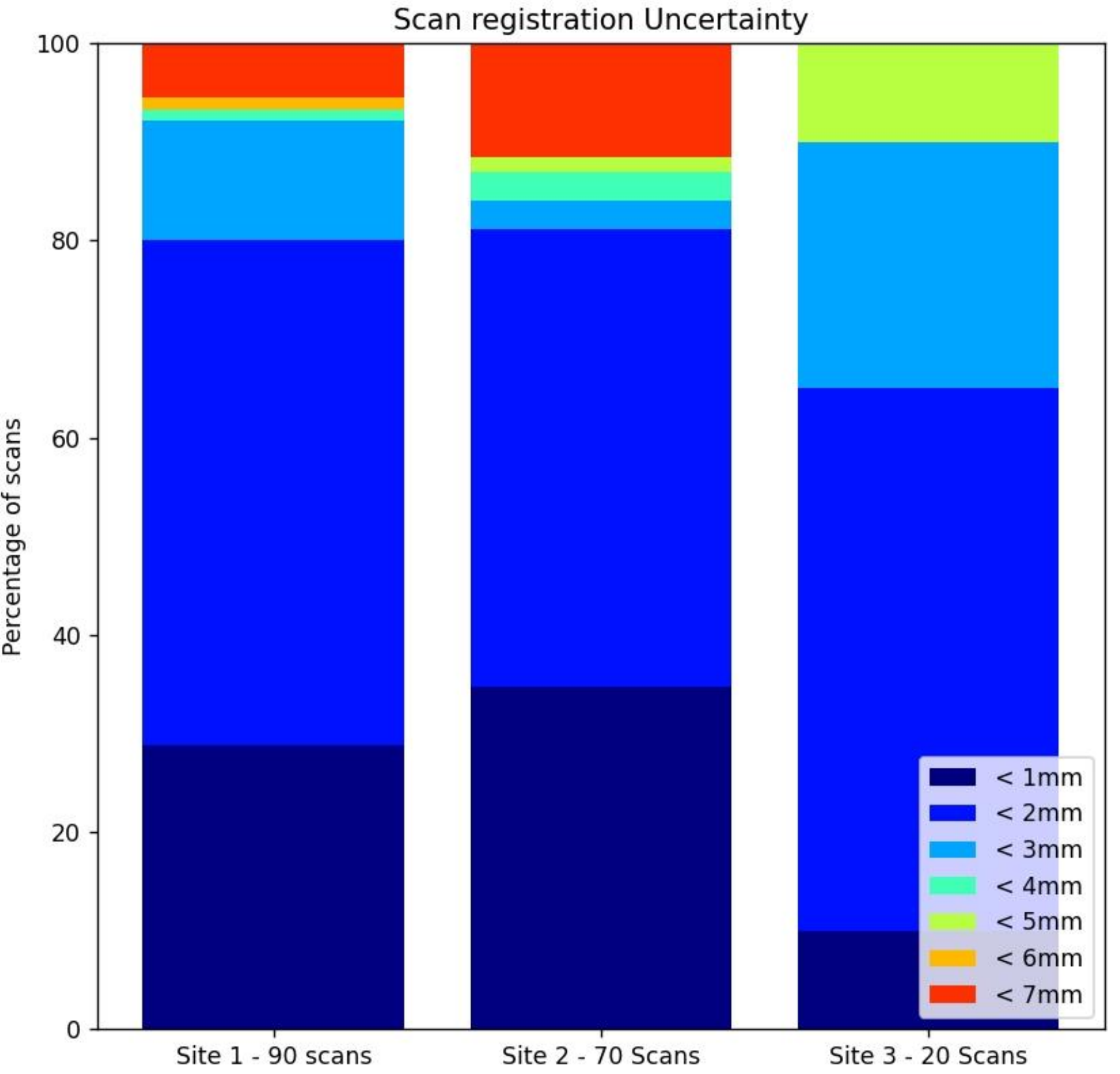}
      \captionsetup{justification=centering}
      \caption{TLS Scan Registration Uncertainty}
      \label{fig:registration}
    \end{subfigure}
    \captionsetup{justification=centering}
    \caption{Dataset Characteristics}
    \label{fig:Sites+TLS+RegUnc}
\end{figure*}

\section{Dataset}

\subsection{Location and Data Overview}
The Hilti SLAM Challenge 2023 dataset features two primary locations (Site 1 and Site 2) and one secondary location (Site 3) as seen in Figure \ref{fig:Site-X}. The reason for having a primary-secondary split is discussed in section \ref{subsec:5D}. Site 1 is a multi-storey new construction with over \SI{4000} {\meter\squared} of floor space, explored via our handheld device, Phasma. It contains sequences with bright to dark transitions, narrow corridors, with multiple areas re-visited. Site 2 is a three-storey underground parking lot under renovation with over \SI{7500} {\meter\squared} of floor space, challenging manhattan world assumptions with non-parallel walls, ramps, and floors containing gradients for storm water drainage. It has sequences from both Trailblazer and Phasma. Site 3 is a tunnel corridor network, parts of which are repurposed as a warehouse. It contains sequences collected exclusively from Phasma, with multiple door transitions, small repetitive structured rooms, and induced system level issues. Note that all sequences were collected within a 48 hour period at each site.

The dataset format has been kept similar to the 2022 Challenge~\cite{Zhang_2023}, with rosbags containing timestamped lidar, cameras and IMU, and GCP name topics. We also provided calibration files and the urdf model for Trailblazer\footnote{\tiny\url{https://github.com/Hilti-Research/trailblazer_description}} in addition to the already public Phasma urdf.

\subsection{Challenge Sequences}
The sequences are designed to have varying degrees of trajectory and map overlap. They are described as follows, along with colored trajectory visualizations in Figure \ref{fig:Site-X}:

\subsubsection{\textbf{Site 1 [Easy-Medium]}}
\begin{itemize}
    \item site1\textunderscore handheld\textunderscore1 (red): The Phasma handheld device makes a loop of the ground floor with multiple re-visits, then going up a staircase to the first floor.
    \item site1\textunderscore handheld\textunderscore2 (green): A traversal of the first floor, then proceeding upstairs towards the second floor where it terminates.
    \item site1\textunderscore handheld\textunderscore3 (blue): A traversal of the second floor, and then up the staircase to the roof, terminating there.
    \item site1\textunderscore handheld\textunderscore4 (white): The sequence initiates near the staircase at the second floor and proceeds downstairs to the basement while exploring areas in each floor en route.
    \item site1\textunderscore handheld\textunderscore5 (yellow): A traversal of the basement level with multiple re-visits, then going up the stairs to the ground floor where it terminates.
\end{itemize}

\subsubsection{\textbf{Site 2 [Easy-Hard]}}
\begin{itemize}
    \item site2\textunderscore robot\textunderscore1 (red): A long spiral loop driven down three storeys of an underground parking renovation site by the Trailblazer robot, with no re-visits.
    \item site2\textunderscore robot\textunderscore2 (green): A loop traversal of a large room at the top level of the garage, as an offshoot of site2\textunderscore robot\textunderscore1. The room has a graded floor.
    \item site2\textunderscore robot\textunderscore3 (blue): This sequence scans a similar large room one level below, with on-robot mounted lighting in a dark environment, and very limited map overlap with site2\textunderscore robot\textunderscore1.
\end{itemize}

\subsubsection{\textbf{Site 3 [Easy-Hard]}}
\begin{itemize}
    \item site3\textunderscore handheld\textunderscore1 (red): A Phasma device traversal from a warehouse area through a door into a small room.
    \item site3\textunderscore handheld\textunderscore2 (green): Multiple door transitions and traversal of a warehouse corridor. It features IMU frame drop instances, and starts in the position where site3\textunderscore handheld\textunderscore1 terminates.
    \item site3\textunderscore handheld\textunderscore3 (blue): Traversal through a U-shaped warehouse corridor - featuring IMU frame drop instances.
    \item site3\textunderscore handheld\textunderscore4 (yellow): Traversal of a part of site3\textunderscore handheld\textunderscore3 in the opposite direction, then a narrow corridor terminating at a section of site3\textunderscore handheld\textunderscore2.
\end{itemize}

\subsection{Additional Sequences}
In Site 2, additional un-evaluated staircase sequences from Phasma were provided to aid with multi-session SLAM since the robot sequences have little to no overlap, thus providing an option for participants to explore multi-device SLAM:
\begin{itemize}
  \item site2\textunderscore handheld\textunderscore4 (yellow): A staircase that connects multi floor regions mapped by site2\textunderscore robot\textunderscore1, that overlap minimally with maps of the large room sequences.
  \item site2\textunderscore handheld\textunderscore5 (white): A staircase that connects multi-floor areas explored in site2\textunderscore robot\textunderscore1, which terminates in a decommissioned bank vault.
  \item site2\textunderscore handheld\textunderscore6 (purple): A staircase sequence that connects the two large rooms.
\end{itemize}

\subsection{Ground Truth Extraction} \label{sec:GT}
The ground truth acquisition methodology was maintained the same as the 2022 Challenge's~\cite{Zhang_2023} approach to sparse ground truth extraction, which leveraged a Terrestrial Laser Scanner (TLS). The primary change was the scanner model - a Trimble X7 TLS (Figure \ref{fig:TLS-x7}) was used instead of the Z\&F Imager 5016. This device change decision is due to the auto registration capabilities of the X7's scanning software that enables this in the field during the data collection process. This greatly reduced the need for manual registration, shortening the processing time by many hours. Trimble's field software also allows for semi-automated survey marker detection, where the user defines the search space for the marker detection system. 92\%, 84\%, and 90\% of registered scans are within 3mm of registration uncertainty or less on a per site basis, as seen in Figure \ref{fig:registration}. Scans containing higher uncertainty were leaf scans containing relatively few connections in the bundle adjustment refinement graph - none of the Challenge's GCPs were extracted from those scans.

\subsection{Privacy Preservation - Face and Licence Plate Blurring}
In order to protect privacy we decided to blur faces of passers-by and car license plates from all the necessary camera streams. For face detection, we leveraged the InsightFace\footnote{\tiny\url{https://github.com/deepinsight/insightface}} library integration for PyTorch. The model deployed was RetinaFace~\cite{deng2019retinaface}, one of the top performers for various face detection challenges. For license plate detection, we used the YoloV5~\cite{ultralytics2021yolov5} model. Hilti employees provided their consent to be recorded. Hence the blurring measure was not taken in some scenarios to minimize processing time. In instances where the blurring system did not function as intended, we created an internal tool for manual blurring to mitigate this at the per-frame, per-device level.

\section{Benchmarking and Evaluation System} \label{sec:benchmarking}
\subsection{Scoring System Update - Increased Accuracy Range}
SLAM frameworks have been improving steadily, as we noticed in the last two years Challenges. Therefore we decided to add a higher accuracy bracket, in order to allocate higher scores for systems that can achieve accuracy within 0.5cm. We also decided to add a bracket to allocate 1 point to trajectory estimates falling within the 10cm to sub-40cm bracket. This was done specifically bearing in mind vision driven systems, as well as multi-session systems. For every GCP evaluated on a trajectory, the score $s_i$ was defined as follows:

\begin{small}
\begin{equation}
  s_{i} =
    \begin{cases}
      20 & \text{if $e_i < 0.005m$} \\
      10 & \text{if $0.005m \leq e_i < 0.01m$} \\
      6 & \text{if $0.01m \leq e_i < 0.03m$}\\
      5 & \text{if $0.03m \leq e_i < 0.06m$}\\
      3 & \text{if $0.06m \leq e_i < 0.1m$}\\
      1 & \text{if $0.1m \leq e_i < 0.4m$}\\
      0 & \text{if $e_i \geq 0.4m$}
    \end{cases}  
\end{equation}
\end{small}

where $e_i$ is the absolute distance error at the $i$th ground control point on that trajectory. The total score for each dataset $S_j$ is the percentage of the maximum possible score (i.e. if all points had \textless0.5 cm error and scored 20 points), except for one dataset that was added later, as explained in a section \ref{subsec:5D}.

\begin{small}
\begin{equation}
    S_j = \left(\dfrac{1}{20N}\sum_{i=0}^{N}{s_i}\right) \times 100
\end{equation}
\end{small}

where N is the number of ground truth points evaluated in each dataset. This denominator normalizes the score for a particular run to be between 0 and 100, regardless of the number of GCP's evaluated in each dataset. The final score is then a sum of the scores from each sequence.

\subsection{Participant Calibration Data Inclusion}
Since the GCP is observed by Trailblazer in the Lidar frame and participant trajectory outputs are in the IMU frame, we provided participants with the option to submit their own extrinsic calibration between LiDAR and IMU in the MultiCal format~\cite{9982031}, for ATE calculation. For teams that did not perform their own calibration, we used our own internally calibrated parameter values.

\subsection{Improved Onboarding: Debug Prompt Notifications}
In the previous Challenge, when participants submitted their solutions to our automatic evaluation server, the system generated an email notifying them of their score. The 2023 Challenge added additional debug functionality to that notification email for teams facing file parsing and formatting errors, invalid quaternions, as well as incomplete trajectories.

\subsection{System Vulnerabilities and Mitigation Measures} \label{subsec:5D}
During the challenge we found that it's possible for teams to experimentally estimate the exact coordinates of the ground control points via a simple euclidean distance formula via the error graph analysis plots that we provide as feedback, and the timestamps at which the system is positioned at a GCP. Manually adjusting the trajectories to achieve better results is not in the spirit of the challenge. Hence we introduced safeguards to mitigate this.
We manually identified the artificially created sparse trajectories, and updated the evaluation system to reject them. We also developed an internal tool to analyze a team's progressive trajectory changes per sequence, enabling observation of manual intervention on the trajectories specifically around the GCP's timestamp by searching for trajectory discontinuities. Accordingly, teams were contacted for clarification and certain submissions were removed. To offset effects of any future occurrences we introduced Site 3 as a secondary location containing exclusively Phasma sequences - having no error analysis plots, and double the maximum achievable score per sequence (200 points). The doubling of scores on Site 3's trajectories would ensure Site 3's maximum achievable score equals the sum of maximum achievable scores from Sites 1 and 2 combined - thereby neutralizing the effects of trajectory manipulation. Formula $(2)$ was hence modified for this location as follows:

\begin{small}
\begin{equation}
    S_3 = \left(\dfrac{1}{20N}\sum_{i=0}^{N}{s_i}\right) \times 200
\end{equation}
\end{small}

\subsection{Extension Towards Multi-session SLAM}
New leaderboard tracks were introduced for vision- and lidar-driven multi-session SLAM evaluation. Teams that opted for it submitted all their trajectories in a common reference frame. The scoring brackets and normalization of total scores remain the same as in single-session. The subtle difference is that in multi-session evaluation, all trajectories for each particular location are considered as one complete trajectory and normalized as such.

\begin{table*}[t]
  \centering

  \resizebox{\linewidth}{!}{%
    \begin{threeparttable}
    \begin{tabular}{|l l|c c c|c c|c c c|c|c c c|}
    \hline
    \multicolumn{14}{| c |}{\textbf{Lidar-Based System Results}} \\
      \hline
      \textbf{Organization} & \textbf{Algorithm} & \multicolumn{3}{ c |}{\textbf{Sensors}} & \multicolumn{2}{ c |}{\textbf{Odometry}} & \multicolumn{3}{ c |}{\textbf{SLAM Backend}} & \textbf{Const.} & \multicolumn{3}{ c |}{\textbf{Results}} \\
       &  & \textbf{Lidar} & \textbf{IMU} & \textbf{Cam}& \textbf{Type} & \textbf{Real-Time} & \textbf{Global BA} & \textbf{Causal} & \textbf{LC} & \textbf{Tuning} & \textbf{GCP Coverage} & \textbf{RMSE ATE(m)} & \textbf{Score} \\
      \hline
      1. KAIST URL & Based on~\cite{lim2023adalio,lim2022single,Segal-RSS-09,frank_dellaert_2022_7383072} & \cmark & \cmark &  & Filter & \cmark & \cmark & \xmark & \cmark & \xmark & 100\% & 0.033 & 1177.64 \\
      2. HKU-MaRS & Based on~\cite{xu2021fastlio2,yuan2022efficient,liu2022efficient,10024300} & \cmark & \cmark &  & Filter+Opt & \cmark & \cmark & \xmark & \xmark & \cmark & 81.82\% & *0.012 & 934.40 \\
      3. Innopolis Univ. & Strelka (Not published) & \cmark & \cmark &  & Filter & \cmark & \cmark & \xmark & \cmark & \xmark & 89.09\% & *0.016 & 840.35 \\
      4. SNU RPM Lab & Based on~\cite{https://doi.org/10.1002/aisy.202200459,Segal-RSS-09,frank_dellaert_2022_7383072} & \cmark & \cmark &  & Filter & \xmark & \cmark & \xmark & \cmark & \xmark & 98.18\% & \textsuperscript{+}0.059 & 789.48 \\
      5. Tsinghua Univ. & FT-LVIO~\cite{https://doi.org/10.1049/rsn2.12376} & \cmark & \cmark & \cmark(1) & Filter & \cmark & \xmark & \cmark & \xmark & \xmark & 100\% & 0.047 & 727.35 \\
      6. ANYbotics & FrankenPharos (Not published) & \cmark & \cmark &  & SW Opt & \cmark & \xmark & \cmark & \xmark & \cmark & 100\% & 0.058 & 657.61 \\
      7. B. Kim et. al. & Based on~\cite{xu2021fastlio2,lim2022single,9560835,frank_dellaert_2022_7383072} & \cmark & \cmark &  & Filter & \cmark & \cmark & \xmark & \cmark & \cmark & 100\% & 0.048 & 629.30 \\
      8. NTU IOT & Based on~\cite{xu2021fastlio2,liu2022efficient,dellenbach2021cticp} & \cmark & \cmark &  & Filter & \cmark & \cmark & \xmark & \xmark & \xmark & 100\% & 0.067 & 597.65 \\
      \hline
      \hline
      \multicolumn{14}{| c |}{\textbf{Vision-Based System Results}} \\
      \hline
      1. Tencent XR & MAVIS~\cite{wang2023mavis} &  & \cmark & \cmark & SW Opt & \cmark & \cmark & \xmark & \cmark & \cmark & 87.27\% & 0.051 & 452.20 \\
      2. KAIST URL & Stereo UV SLAM~\cite{lim2022uv}  &  & \cmark & \cmark & SW Opt & \cmark & \cmark & \xmark & \cmark & \cmark & 78.18\% & \textsuperscript{+}0.190 & 200.80 \\ 
      3. ASL ETHZ &  Based on~\cite{20.500.11850/100060,cramariuc2022maplab} &  & \cmark & \cmark & SW Opt & \cmark & \cmark & \xmark & \cmark & \xmark & 100\% & 0.189 & 121.20 \\
      \hline
    \end{tabular}
    \begin{tablenotes}
        \item[*: RMSE ATE is only calculated for data that is submitted. Cases where teams receive lower/better RMSE ATE numbers but poor ranking is as an indicator that sequence trajectories are either incomplete or skipped]
        \item [\textsuperscript{+}: This team receives a higher score than subsequent leaderboard entries despite lower GCP coverage and higher RMSE ATE, due to its best GCP estimates being in a higher scoring band.]
    \end{tablenotes}
    \end{threeparttable}
  }
  \captionsetup{justification=centering}
  \caption{Single-session SLAM leaderboards - Top entries}
  \label{table:1}
\end{table*}

\begin{table*}[t]
  \centering

  \resizebox{0.6\linewidth}{!}{%
    \begin{threeparttable}
    \begin{tabular}{|l l|c c c|c|c c c|}
    \hline
    \multicolumn{9}{| c |}{\textbf{Lidar-Based System Results}} \\
      \hline
      \textbf{Organization} & \textbf{Algorithm} & \multicolumn{3}{ c |}{\textbf{Sensors}} & \textbf{Fully} & \multicolumn{3}{ c |}{\textbf{Results}} \\
       &  & \textbf{Lidar} & \textbf{IMU} & \textbf{Cam} & \textbf{Automated} & \textbf{GCP Coverage} & \textbf{RMSE ATE} & \textbf{Score} \\
      \hline
      1. ETHZ ASL & Based on~\cite{xu2021fastlio2,cramariuc2022maplab} & \cmark & \cmark & \cmark & \cmark & 100\% & 0.294 & 36.7 \\
      2. Innopolis Univ & Strelka (Not published) & \cmark & \cmark &  & \xmark & 12.72\% & 0.359 & 1.7 \\
  
      \hline
      \hline
      \multicolumn{9}{| c |}{\textbf{Vision-Based System Results}} \\
      \hline
      1. Tencent XR & MAVIS~\cite{wang2023mavis} &  & \cmark & \cmark & \cmark & 83.63\% & 0.287 & 27.0 \\
      2. ETHZ ASL & Based on~\cite{20.500.11850/100060,cramariuc2022maplab} &  & \cmark & \cmark & \cmark & 100\% & 0.397 & 15.3 \\ 
      \hline
    \end{tabular}
    \end{threeparttable}
  }
  \captionsetup{justification=centering}
  \caption{Multi-session SLAM leaderboards}
  \label{table:2}
\end{table*}

\section{The 2023 Challenge}
\subsection{Overview}
The Hilti SLAM Challenge 2023 saw an increase in participation over the previous iterations with 69 unique teams participating, compared to 42 from 2022, and 27 from 2021. This indicates an increasing level of trust being placed in the dataset and benchmark by the worldwide SLAM community. Results were announced at the \textit{IEEE ICRA 2023 2nd workshop on the Future of Construction}. Our live leaderboard contains an even higher participant count after the Challenge completed, signalling continued interest.

\subsection{Approaches of Winning Systems}
The Lidar driven single-session category in Table \ref{table:1} was won by the Urban Robotics Lab at KAIST, which leveraged AdaLIO~\cite{lim2023adalio} as its SLAM frontend. The backend used Quatro~\cite{lim2022single} to detect loop closures, after which pose graph optization was performed via a factor graph. For Sites 2 and 3, this system was extended with multi-session map-to-map G-ICP~\cite{Segal-RSS-09} which also included additional handheld sequences, to improve the single-session maps. They however did not participate in the multi-session track. Furthermore, specifically for site2\textunderscore robot\textunderscore 1, GTSAM~\cite{frank_dellaert_2022_7383072}'s OrientedPlane3Factor was used in the backend, as seen in their report\footnote{\tiny\url{https://submit.hilti-challenge.com/submission/1edf81f7-b147-64c2-9787-4771ad2a8f7e/report}}, leveraging an Atlanta world assumption to minimize drift.

The category winner for LiDAR-driven multi-session SLAM was the Autonomous Systems Lab at ETH Zurich. The system leveraged Maplab~\cite{cramariuc2022maplab} with lidar, vision, and IMU sensor modalities. LiDAR odometry was conducted with FastLio2~\cite{xu2021fastlio2}.
Maplab-compatible SuperPoint~\cite{detone2018superpoint} and SuperGlue~\cite{sarlin20superglue} features were used for feature detection and tracking, followed by per-trajectory bundle adjustment. Loop closures were searched using HNSW~\cite{malkov2016efficient} and corresponding landmarks were merged. The map was reoptimized with these new constraints, and BALM~\cite{liu2022efficient} was also integrated.
Multi-session mapping was done by rigidly aligning maps. All frames were matched, a set of transformations computed, and RANSAC was used to find the best alignment.
Then, global bundle adjustment and loop closure were run for a few iterations, followed by re-projection error based landmark rejection, while adjusting camera intrinsics for improved accuracy.

The category leader for vision-driven SLAM in both single and multi-session tracks was Tencent Games' XR Vision Lab with their TXR-SLAM\footnote{\tiny\url{https://submit.hilti-challenge.com/submission/1edf5fd1-fff6-6e58-9f18-c7f17ae082ca/report}} system, later revealed as MAVIS: Multi-Camera Augmented Visual-Inertial SLAM~\cite{wang2023mavis} after it was published. Their core contribution is a novel, exact IMU pre-integration formulation based on the exponential function of an automorphism of the SE\textsubscript{2}(3) matrix Lie group. The backend is optimization-based, containing sliding window bundle adjustment and leveraging global bundle adjustment at loop closure instances. For multi-session SLAM, a bag-of-words model is leveraged. None of the additional sequences were used for multi-session alignment. Also, front cameras were not used because lidar lines were visible. Site 3's IMU frame drop instances were interpolated via a cubic spline.

\subsection{Leaderboard Trends and Takeaways}
The 2023 Challenge witnessed multiple interesting trends. The Lidar single session category is led by academia, with KAIST securing the top spot. Hierarchical Bundle adjustment \cite{10024300} is a new high-performing system backend created by runner-up HKU-MARS Lab. They also skipped Loop Closure altogether. There are also high-performing open-loop odometry-only systems in the top teams such as Tsinghua University's FT-LVIO~\cite{https://doi.org/10.1049/rsn2.12376} system and ANYBotics' FrankenPharos\footnote{\tiny\url{https://submit.hilti-challenge.com/submission/1edf8227-fc31-6db8-bbd5-c34ba577a9cc/report}}, which secured the 5th and 6th spots on the leaderboard. Very few camera-augmented lidar SLAM systems are in the top 10 teams, with FT-LVIO ~\cite{https://doi.org/10.1049/rsn2.12376} being the only entry of that kind. The Vision Single Session Category is led by big-tech, with Tencent Games' XR Lab with their MAVIS system securing first place. The category also finds all top SLAM frontends being optimizer based. On comparing Lidar-based SLAM vs. Vision-based SLAM, the RMSE ATE accuracy gap appears to be closing when compared to the 2022 Challenge leaderboard. Interestingly, none of the top teams in the displayed leaderboards included their own extrinsic calibration in their final submissions. On average, robot trajectories resulted in few centimeters higher error than handheld trajectories, likely due to 3x higher lidar noise and non-Manhattan spaces.

The Multisession SLAM tracks (Table \ref{table:2}) for lidar and vision-based systems saw relatively limited participation, with two teams participating in each track, which was anticipated given the limited number of multi-session systems available. Both vision- and lidar-driven multi-session systems received similar scores at similar accuracies in the 30cm range of RMSE ATE. But this changed as of late 2023 when multiple new systems populated the lidar multi-session leaderboard\footnote{\tiny\url{https://www.hilti-challenge.com/leader-board-2023.html}}. Limited map overlap between site2\textunderscore robot\textunderscore 1 and site2\textunderscore robot\textunderscore 3 was a corner case where multi-device integration was the only mitigation measure. No teams used timestamped GCP names or relative positions in establishing cross-trajectory constraints.

\subsection{Known Issues}
Our ground truth system is based on the sparse ground truth extracted from a unified terrestrial laser scan on a per-site basis, without dense trajectory coverage. Regarding calibration, we acknowledge that our system does not have an external validation methodology for accuracy. Hence, we released calibration datasets and provided teams with the option to apply their own extrinsic calibration for evaluation. Lastly, in our robot calibration, the rotation of the roll and pitch axes has been limited to the amount of flex that the manual pallet jack could tolerate.

\section{Conclusions}
With the 2023 SLAM Challenge iteration, we are delighted to extend our dataset and benchmarking abilities into multi-device and multi-session SLAM. We witness top-performing single-session lidar SLAM systems attain sub 2cm accuracy in general case handheld scenarios. We also see the accuracy gap between vision- and lidar-driven SLAM reducing. Live leaderboards continue to be available for public use. We hope that the SLAM systems using our dataset and benchmark are deployed to benefit mankind.

\section*{Acknowledgments}
 We would like to thank all the SLAM teams worldwide for their continued participation and engagement, and Hilti team members both current and former - Michael Helmberger, Nadine Imholz, Alejandro Marzinotto, Lukas Blass, Mathis Dumas, and Kristian Morin, for their contributions.


\bibliographystyle{ieeetr}
\bibliography{refs}

\end{document}